\documentclass[10pt,twocolumn,letterpaper]{article}

\usepackage{cvpr}
\usepackage{times}
\usepackage{epsfig}
\usepackage{graphicx}
\usepackage{amsmath}
\usepackage{amssymb}

\DeclareMathOperator*{\argmax}{arg\,max}
\DeclareMathOperator*{\argmin}{arg\,min}

\usepackage[pagebackref=true,breaklinks=true,letterpaper=true,colorlinks,bookmarks=false]{hyperref}

\cvprfinalcopy 

\def\cvprPaperID{****} 

\ifcvprfinal\fi
\begin{document}

\title{Fine-Grain Few-Shot Vision via Domain Knowledge as Hyperspherical Priors}

\author{Bijan Haney\\
Augustus Intelligence\\
{\tt\small bijan.haney@augustusai.com}
\and
Alexander Lavin\\
Augustus Intelligence\\
{\tt\small alexander.lavin@augustusai.com}
}

\maketitle

\begin{abstract}
Prototypical networks have been shown to perform well at few-shot learning tasks in computer vision. Yet these networks struggle when classes are very similar to each other (fine-grain classification) and currently have no way of taking into account prior knowledge (through the use of tabular data). Using a spherical latent space to encode prototypes, we can achieve few-shot fine-grain classification by maximally separating the classes while incorporating domain knowledge as informative priors. We describe how to construct a hypersphere of prototypes that embed \textit{a priori} domain information, and demonstrate the effectiveness of the approach on challenging benchmark datasets for fine-grain classification, with top results for one-shot classification and 5x speedups in training time.
\end{abstract}

\section{Introduction}

Large-scale image classification datasets such as ImageNet are not representative of most real-world scenarios: datasets are often small, sparsely labeled if at all, and with imbalanced distribution of classes. Research towards few-shot learning algorithms aims to address these concerns. \textit{Few-shot} is recognizing concepts from small labeled sets, where typically a model trains on labeled data of base classes and can classify unseen novel classes using only a few examples \cite{Miller2000LearningFO,Li2006OneshotLO,Vinyals2016MatchingNF}. \textit{Prototypical networks}, which aim to match examples with nearest class prototypes, have some of the highest few-shot accuracies \cite{Snell2017ProtoFewShot}. Sophisticated regularization techniques like Manifold Mixup and additional auxiliary tasks can further improve performance \cite{Mangla2019ManifoldMixup}. We refer the reader to \cite{Wang2019GeneralizingFA} for a recent survey of few-shot methods.

Yet these networks, and mainstream convolution neural network (CNN) methods for that matter, struggle when only small interclass differences exist and specific details matter. This is the task of \textit{fine-grain} classification, where even human annotators struggle. Effective methods use CNNs with innovative learning approaches such as multi-scale \cite{Zhang2020ThreebranchAM}, transfer \cite{Kolesnikov2019LargeSL}, and curriculum \cite{Chen2019DestructionAC} learning.

The combination of few-shot and fine-grain is nefarious, yet accounts for the majority of real-world use of computer vision -- for example, defect classification in manufacturing or machinery inspection, or microscopic examination of tissue to study the manifestations of disease (histopathology).

\begin{figure}[!tb]
\begin{center}
\fbox{
  \includegraphics[width=0.95\linewidth]{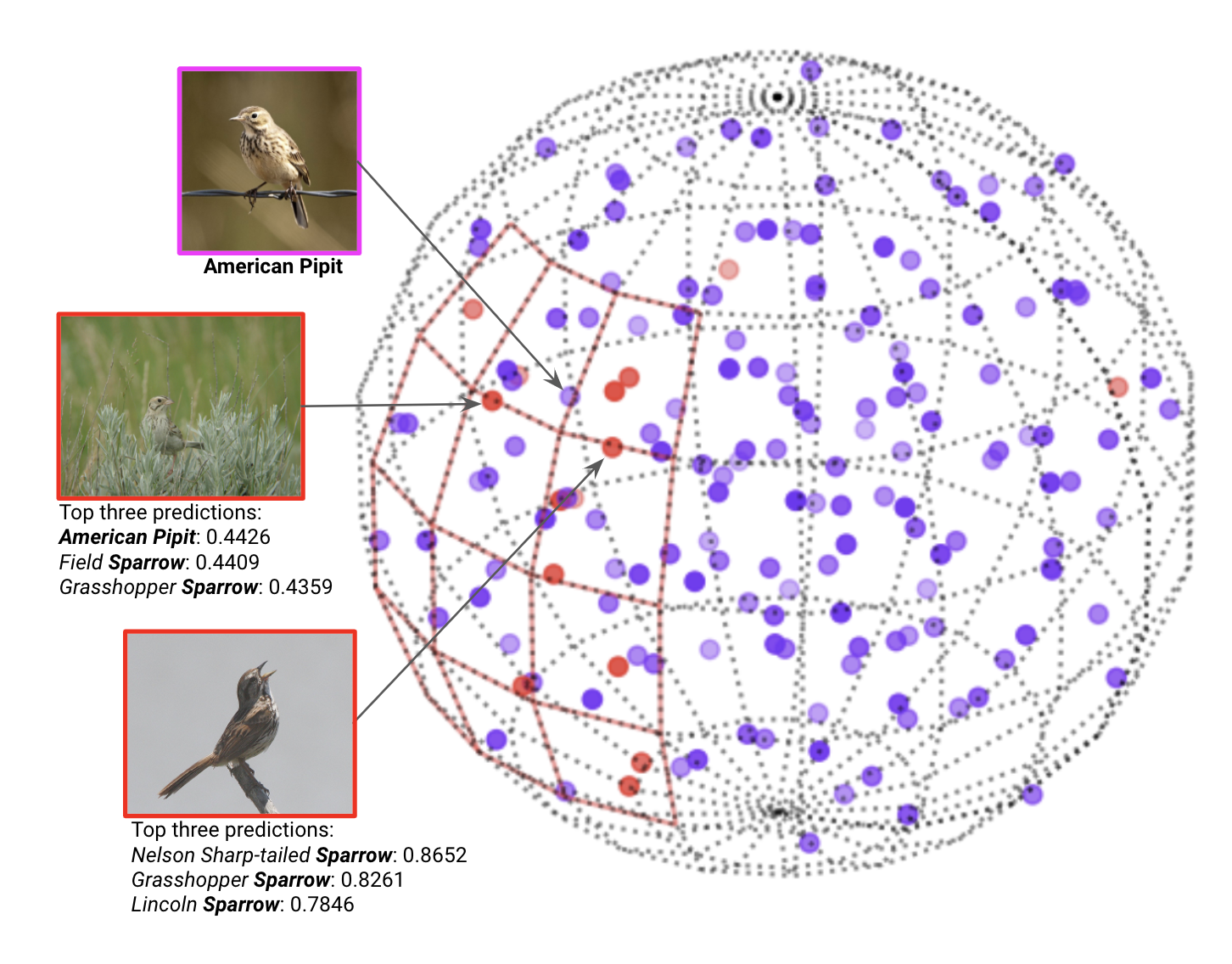}
   }
\end{center}
   \caption{Because we encode a hyperspherical latent prototype space with bird species taxonomy info, inference with a never-before-seen sparrow class gets very close to the other sparrow classes (red region). This is true for both the clear (bottom) and noisy (middle) \textit{Baird Sparrow} instances. The latter also infers with low confidence an \textit{American Pipet} (top), nearby in semantic space.
   }
\label{fig1}
\end{figure}

To address these challenges we suggest utilizing available domain knowledge to define class prototypes \textit{a priori}, embedded in a hyperspherical latent space \cite{Mettes2019HypersphericalPN}. This modeling structure can massively bootstrap learning and yield more precise classification.
Our contributions:
\begin{enumerate}
    \item Theoretical utilities of non-Euclidean latent spaces towards few-shot learning and fine-grain classification in computer vision.
    \item Proposal of hyperspherical prototypes derived from \textit{a priori} domain knowledge (building off of \cite{Mettes2019HypersphericalPN}).
    \item Empirical analyses of our novel prototype network model towards few-shot classification on a challenging real-world fine-grain dataset of bird species identification.
\end{enumerate}

\section{Hyperspherical Prototypes}

Prototype-based networks for classification employ a metric output space and divide the space into Voronoi cells around a prototype per class, typically the mean location of training examples. Intuitively this representation lends itself to the few-shot problem, where the task simplifies to matching new class instances to their nearest prototype. Classic approaches use a Euclidean distance metric, but this is shown to scale poorly to high dimension embedding spaces \cite{Lee2019MetaLearningWD}. The choice of latent space can help maximally spread the prototypes, and non-Euclidean spaces can allow for smooth interpolation.

Hyperspheres are bounded in space $[-1,1]^D$, while a Euclidean space is unbounded $(-\infty,\infty)^D$. If the latent features are meant to represent visible attributes of the class that are either present or not, then a bounded space is a better representation of the binary latent features. Another advantage of a spherical space is for clustering  examples of the same class together when they have large intraclass differences. Projections should be insensitive to the magnitude of the feature, while sensitive to its presence, and thus there is an advantage to have all examples live on the surface of the hypersphere.

Recently \cite{Mettes2019HypersphericalPN} proposed defining class prototypes \textit{a priori} with maximal separation on a hypersphere. The main advantage being data-independent optimization, so new data instances don't require prototype updates. The authors suggest the optimal set of prototypes, $\mathbf{P}^{\ast}$, from the space $\mathbb{P}$ of $a$-dimensional vectors with magnitude $1$, is the one where the largest cosine similarity\footnote{The hypersphere is a valid metric space and thus allows us to compute distances between any pair of points. The natural pairwise (dis)similarity metric to use is cosine distance.} between two class prototypes $\mathbf{p}_i$, $\mathbf{p}_j$ 
from the set is minimized:

\begin{equation}
\mathbf{P}^{\ast} 
= \argmin_{\mathbf{P} \in \mathbb{P}} \Big(
    \max_{(m,l,m \neq l) \in C}
    \cos \theta_{\mathbf{p}_m,\mathbf{p}_l}
    \Big)
\end{equation}

where we have $M$ class prototypes and $a$ output dimensions.

\begin{figure}[!tb]
\begin{center}
\fbox{
  \includegraphics[width=0.95\linewidth]{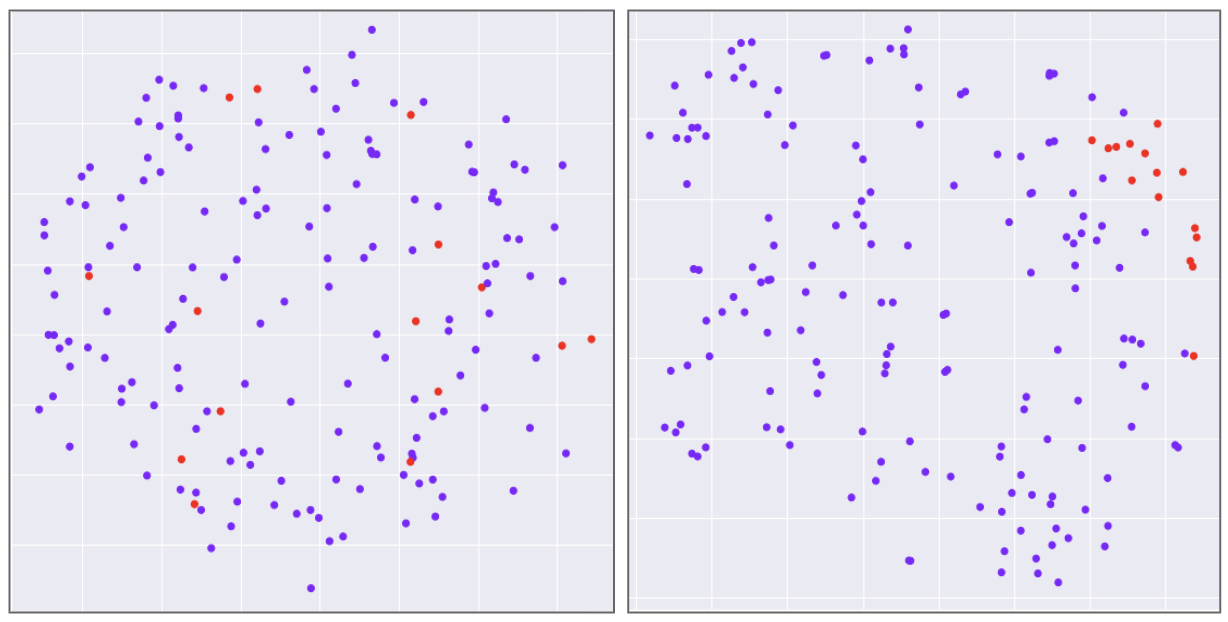}
   }
\end{center}
   \caption{T-SNE projections of class prototypes in hyperspherical space with and without domain priors (left and right, respectively). The former randomly distributes the classes -- there is no proximity with semantically similar classes, e.g. sparrows (red). Our embedding method results in localized prototypes with useful semantics, while maintaining near-maximal spread.
   }
\label{fig2}
\end{figure}

However this embedding treats classes independently and is thus not a well-regularized metric space. As shown in Fig. \ref{fig2}, different sparrow classes project to dispersed locations in the latent space. This is troublesome for fine-grain classification as the output vector will swing wildly across the latent space with subtle changes in the input vector (e.g., a slight difference in beak length). This is also the case with standard prototypical networks that use a poorly regularized Euclidean latent space.

The $\mathbf{P}^{\ast}$ hypersphere spreading does not make use of available prior information of classes. \cite{Mettes2019HypersphericalPN} suggest word embeddings of class names could provide semantic information. While these come for free, they can be misleading and actually add noise in the case of fine-grain classification. For example, in classifying bird species there is effectively no semantic difference between the class names ``Florida jay" and ``cardinal", yet these species are easy to distinguish based on actual features (colored blue and red, respectively). And consider many classes have names like ``cedar waxwing" and ``brewer blackbird" that have words with non-bird semantics.

\subsection{Encoding Domain Knowledge as Class Priors}

\begin{table}
\begin{center}
\resizebox{\columnwidth}{!}{
\begin{tabular}{|c|c|c|c|c|c|}
\hline
Img Id & class & back color: & ... & wing color: & wing pattern:\\
      &       & black       &      &  yellow     & spotted\\
\hline\hline
1 & Black footed Albatross & 0 & ... & -1  & 0\\
2 & Black footed Albatross & -1 & ... & -1  & 0\\
... & ... & ... & ... & ...  & ...\\
346 & Parakeet Auklet & 1 & ... & -1  & -1\\
347 & Parakeet Auklet & 1 & ... & -1  & -1\\
\hline
\end{tabular}
}
\end{center}
\caption{Example of tabular data associated with the images from the CUB training data set. Each row is a set of visual features that describe the bird in the image, such as whether it has a hooked bill shape, brown wing color, etc. There are $A = 312$ features in the CUB tabular data set. Presence of the attribute in the image is denoted with $+1$, absence $-1$, and $0$ if it could not be determined. }
\label{tabledata}
\end{table}

In many real-world domains 
such as medicine (e.g. tissue and joint analysis), predictive maintenance (e.g. steel grades and properties), and biology (e.g. species identification or plant pathology) 
there is rich domain knowledge to utilize that can massively bootstrap learning and improve precision.

Consider we are given $N$ training examples with both tabular and image features $\{(x_i,w_i,y_i)\}^N_{i=1}$, where $x_i \in R^A$ is an $A$-dimensional tabular input,  $w_i \in R^V$ is an $V = R^{H \times W \times C}$ 2D image input with $H \times W$ pixels and $C$ channels, and $y_i \in K$ is the class label of the $i$-th training example, where $K=\{1,..,M\}$ is the set of $M$ class labels.

We consider a model, $f_{\phi}(\mathbf{w})$, which takes as input an image $w_i$ and has hidden, trainable weights $\phi$. The model $f_{\phi}(\mathbf{w})$ projects the input image to the surface of a hypersphere of dimension $a \leq A$, i.e. 

\begin{equation}
f_{\phi}: R^{V} \rightarrow \mathcal{S}^{a-1}
\end{equation}

$f_{\phi}(\mathbf{w})$ is trained by a loss function aiming to maximize the cosine similarity between the output vector and a fixed class prototype vector. 

The the latent features of the hyperspherical space encode domain knowledge by using the training set tabular data $T = \{(x_i,y_i)\}^N_{i=1}$ as prior information. An example of tabular data is given in Table \ref{tabledata}. The data set $T$ is first transformed to have a dimensionality of $a$, the dimensionality of the latent space. $a$ is a hyperparameter of the model.  

In order to choose which $a$ transformed features from the $A$ original features will be used in the latent space, we construct a decision tree to identify the most informative features as determined by which features offer the largest reduction in entropy at each node of the tree. We then reduce the dimensionality of the tabular dataset $T$ such that each sample $x_i \in R^a$. 

After the dataset is reduced, for each class in $K$, we average all training examples ${x_i}$ of that class together and normalize the vector to unity. This creates $M$ normalized attribute prior vectors $\mathbf{\alpha}_m \in \mathcal{S}^{a-1}$. 

\begin{equation}
\mathbf{\alpha}_m = \frac{\sum_{i=1}^{N_m} x_i^m}{\|\sum_{i=1}^{N_m} x_i^m\|} \qquad \mathbf{\alpha}_m^T\mathbf{\alpha}_m = 1
\end{equation}

Where $x_i^m$ is a sample for which $y_i = m$ and $N_m$ is the number of samples in class $m$. These attribute vectors can either be used directly as the prototype vectors of the classes in the latent space, or they can be used to label prototype vectors created by another method. 

In the experiments section below we quantify the difference between prototypes created with domain-knowledge and prototypes that are simply maximally spread out in the space. Importantly for few-shot, the inference of never-before-seen classes does not need the prior attribute tabular data used in training the other classes.

We also investigate the behavior of the latent hypersphere in high dimensions: in the infinite limit of dimensions $a$ the surface area of a hypersphere approaches $0$. Even at $a>20$,  \cite{Davidson2018HypersphericalVA} observe this \textit{vanishing surface problem}. We find this empirically not to be a problem, as shown in the results later (Table \ref{table2}).

Notably recent works with Poincar{\'e} latent spaces \cite{Mathieu2019ContinuousHR,Klimovskaia2019PoincarMF} show this to be a natural  representation for modeling and inferring tree-structured data. That is, hyperbolic geometry is a continuous analogue to trees and enables low-distortion embeddings of hierarchical structures. To get interpretable semantics from the dimensions, use a Poincar{\'e} or other hyperbolic space. Comparatively, spherical is best for class separation. 

\subsection{Model Learning \& Inference}

Given the $M$ \textit{a priori} class average attribute vectors $\alpha_m$ as described in the previous section, one can use the class average attribute vectors directly as class prototypes (i.e. $\mathbf{p}_m = \alpha_m $). Another way to generate the prototypes is to create maximally spaced apart prototype vectors $\mathbf{P}=\{\mathbf{p}_1,...,\mathbf{p}_M\}$ as described in \cite{Mettes2019HypersphericalPN}. Each prototype $\mathbf{p}$ is then assigned a class label based on the its nearest attribute vector $\alpha_m$ as determined by cosine similarity. Once a class is assigned to a prototype, no other prototypes can be assigned that class label. In this way, classes of similar attributes are clustered together while still maintaining maximum distance. 

For training, we use the image dataset of $N$ examples from $M$ classes $\{(w_i,y_i)\}^N_{i=1}$. Our model $f_{\phi}(\mathbf{w})$ is a vanilla CNN architecture, ResNet32. We train the weights $\phi$ of the model to project input images $\mathbf{w}$ to an $a$-dimensional vector in the hyperspherical latent space. The loss function is specific to the hypersphere class embedding space: as shown in \cite{Mettes2019HypersphericalPN} it maximizes the cosine similarity between the output vectors of the training examples and their corresponding class prototypes.\footnote{We note \cite{Gidaris2018DynamicFV} show that using cosine similarity in the output helps generalization to new classes, yet our focus here is on the prototype space.}

For a test image we perform inference by computing the cosine similarity to all class prototypes and we select the class with the highest similarity:

\begin{equation}
k^{\ast} = \argmax_{k \in K} \cos(\theta_{f_{\phi}(\mathbf{w}),\mathbf{p}_c})
\end{equation}

where $f_{\phi}(\mathbf{w})$ is the inferred hypersphere point for test image $\mathbf{w}$.

\section{Experiments}

We evaluate our approach on the CUB-200-2011 dataset \cite{Wah2011TheCB}, an ideal benchmark for fine-grained visual recognition algorithms. It contains 200 different bird species, along with annotations for bird bounding box, 15 part locations, and 312 binary attributes; the attributes we utilize as \textit{a priori} domain information, as described earlier. Only 170 classes are used for training and validation, and 30 classes are held out for the test set. During the testing phase, the original prototypes are discarded, and new prototypes are generated for the unseen test classes, either using one example for the prototype (1-shot) or the average of 5 examples (5-shot) as described in \cite{Snell2017ProtoFewShot}.

\subsection{Implementation Details}
For all our experiments we use a standard ResNet32 backbone with the following setup: SGD optimizer, a learning rate of 0.01, momentum of 0.9, weight decay of 1e-4, batch size of 128, no gradient clipping, and no pre-training. All networks are trained for 250 epochs, an order of magnitude reduction in learning rate at 100 and 200 epochs. For data augmentation, we perform random cropping and random horizontal flips. All runs are repeated with five random seeds and we report the average results.

\subsection{Results}

We compare our hyperspherical domain prototype network to other prototype networks that have proven effective on the few-shot task. Results are shown in Table \ref{table1} for both the one-shot five-way and five-shot five-way tasks. Clearly our method is superior to other prototypical network approaches that use a standard Euclidean latent space, and that use a maximally spread spherical space.

We noted earlier a key advantage of the latent space modeling in \cite{Mettes2019HypersphericalPN} and used here is data-independent optimization: the prototype space doesn't need updating with new data. We observe a near 5x speed-up in training time: 20 seconds per epoch, versus 96 seconds per epoch for the standard Euclidean prototypical network (each with batch size 128, 128 dimensions, on a GeForce RTX 2080 Ti).


\begin{table}
\begin{center}
\begin{tabular}{|c|c c|c|}
\hline
Embedding & One-shot & Five-shot & Train time (s)\\
\hline\hline
Euclidean & 53.91 & 66.49 & 96\\
Hyperspherical & 53.33 & 70.27 & \textbf{20}\\
+ Domain priors & \textbf{59.48} & \textbf{71.19} & \textbf{20} \\
\hline
\end{tabular}
\end{center}
\caption{Percent accuracies on 1- and 5-shot 5-way task with CUB dataset. The first method is a standard prototypical net with Euclidean embedding space, followed by the hyperspherical prototype net \cite{Mettes2019HypersphericalPN} with maximally spread prototypes, and finally our method that includes domain priors on the hypersphere prototypes. The training times are seconds per epoch. See text for details.}
\label{table1}
\end{table}

Note these results reflect a hyperspherical domain prototype network that uses an off-the-shelf ResNet32 CNN. Undoubtedly the performance can be improved with a fine-tuned architecture, which we will explore in forthcoming work. Our method could also be supplemented by the aforementioned multi-scale and curriculum learning approaches. For now those are beyond scope as our focus is on latent embedding spaces.

We mentioned earlier the theoretical bound on surface area for a hypersphere in high dimensions: The maximum surface area exists at $a=7$ dimensions, and \cite{Davidson2018HypersphericalVA} suggest $a>20$ to exhibit a \textit{vanishing surface}. For our network on the one-shot five-way task, we instead observe the results in Table \ref{table2}, and thus no unstable behavior of hyperspherical models in high dimensions.


\begin{table}
\begin{center}
\begin{tabular}{|c|c c c c|}
\hline
Dimensions & 7 & 20 & 64 & 128 \\
\hline
One-shot (acc\%) & 44.34 & 53.43 & 57.23 & 59.48 \\
\hline
\end{tabular}
\end{center}
\caption{Testing a variety of hypersphere dimensions, we do not observe the \textit{vanishing surface problem} from spatial collapse at dimensions $>20$.}
\label{table2}
\end{table}

\section{Conclusion}

This paper demonstrates hyperspherical embeddings of class prototypes towards few-shot fine-grain classifications.
We build on the work of \cite{Mettes2019HypersphericalPN} to realize the potential of this representational choice: encode domain knowledge as informative priors, which can enable computer vision in real-world applications with limited data and the need for precise classification.
In future work we anticipate implementing dual loss functions that maximize separation while maintaining semantic information.

{\small
\bibliographystyle{ieee_fullname}
\bibliography{fgvc20}
}

\end{document}